\documentclass[letterpaper, 10 pt, conference]{ieeeconf}  

\IEEEoverridecommandlockouts                              

\usepackage{newtxmath} 

\usepackage{multirow}
\usepackage[pdftex]{graphicx}
\usepackage{amsmath}
\usepackage{bm}

\title{\LARGE \bf
An Estimation Framework for Passerby Engagement \\ Interacting with Social Robots
}

\author{Taichi Sakaguchi$^{1}$ and Yuki Okafuji$^{1}$ and Kohei Matsumura$^{1}$ and Jun Baba$^{2,3}$ and Junya Nakanishi$^{2}$
\thanks{$^{1}$
T. Sakaguchi, Y. Okafuji and K. Matsumura are with College of Information
Science and Engineering, Ritsumeikan University, Kusatsu, Shiga, JAPAN.}
\thanks{$^{2}$J. Baba and J. Nakanishi  are with Graduate School of
Engineering Science, Osaka University, Toyonaka, Osaka, Japan.}
\thanks{$^{3}$J. Baba is with CyberAgent Inc., Shibuya Scramble Square 22F,
Shibuya, Tokyo, Japan}
}

\begin{document}

\maketitle
\thispagestyle{empty}
\pagestyle{empty}

\begin{abstract}
Social robots are expected to be a human labor support technology, and one application of them is an advertising medium in public spaces. When social robots provide information, such as recommended shops, adaptive communication according to the user's state is desired. User engagement, which is also defined as the level of interest in the robot, is likely to play an important role in adaptive communication. Therefore, in this paper, we propose a new framework to estimate user engagement. The proposed method focuses on four unsolved open problems: multi-party interactions, process of state change in engagement, difficulty in annotating engagement, and interaction dataset in the real world. The accuracy of the proposed method for estimating engagement was evaluated using interaction duration. The results show that the interaction duration can be accurately estimated by considering the influence of the behaviors of other people; this also implies that the proposed model accurately estimates the level of engagement during interaction with the robot.

\end{abstract}

\section{INTRODUCTION}
In recent years, a variety of social robots have been developed and deployed in public spaces, such as shopping malls. For instance, they are used for advertising mediums, such as guiding users around the facility \cite{gross2008guide} and distributing flyers \cite{shi2018robot}. When social robots provide information to people as the advertising medium in public spaces, they require adaptive communication. For instance, if a low-engagement user is talking to a robot, the robot should ask, "Are you listening?" Therefore, user engagement, which is also defined as the level of interest in the robot, is likely to play an important role in implementing adaptive communication. Thus, some studies have proposed methods for estimating users engagement \cite{xu2013designing, bohus2014managing, del2020you, leite2015comparing, salam2015engagement, ben2019fly, inoue2019latent}.

Xu et al. proposed an estimation method for engagement of user communicating with the robot that uses support vector machine (SVM) with the user’s posture, gaze direction, and the distance between the user and the robot \cite{xu2013designing}. Bohus and Horvitz estimated user's engagement using a regression model with user's body features \cite{bohus2014managing}. Del Duchetto et al. proposed an End-to-End method which estimates the user engagement from camera images \cite{del2020you}. These estimation methods of user engagement represent the level of user interest in the robots. These studies aim to apply the estimation method to the robot's behavioral strategy to design flexible interactions, such as changing talk contents.

While various estimation methods of engagement have been proposed, Oteral et al. point out eight unsolved open problems regarding engagement estimation \cite{oertel2020engagement}. Among them, in this paper, we focus on four unsolved points: A) Multi-party interactions, B) Process of state change in engagement, C) Difficulty in annotating engagement, and D) Interaction dataset in the real world. By proposing a new estimation framework for solving these four open problems, this paper will provide new insights into user engagement estimation methods.

Regarding the four open questions that this study focused on, there are several studies that try to solve each individual problem. The outline of each problem is shown below.

\textbf{A) Multi-party interactions:} Few studies consider the influence of multi-party interaction for estimating engagement \cite{leite2015comparing, salam2015engagement}. Leite et al. estimated the user engagement using SVM with some features, such as body gestures. They compared two engagement estimation models applied to participants in multi-party interaction: a model trained on data from individual participants, and a model trained on data from multi-party interactions \cite{leite2015comparing}. Slam and Chetouani proposed an engagement estimation method using SVM with body features obtained from another user in a situation where two users interact with one robot \cite{salam2015engagement}. Both studies show that user engagement can be accurately estimated by considering the features of others in the estimation model. This implies that the engagement of the user during the interaction with the robot is influenced by others. However, these previous studies have a problem in that the number of other users during the interaction is fixed. Group sizes in real-world interactions vary; thus, an estimation method that can consider various group sizes is required.

\textbf{B) Process of state change in engagement:} Numerous studies related to estimating engagement deal with engagement as a binary value: whether or not the user is interested in the robot \cite{leite2015comparing, salam2015engagement}. By assuming the binary value in this manner, it can be easily estimated by a clustering method, such as SVM, whereas user engagement is considered a continuously changing process. If we can estimate the process of state change in engagement, it can be applied to predict the time when users terminate the interaction with the robot. Ben-Youssef et al. proposed a method that predicts a decline in engagement \cite{ben2019fly}. However, to the best of our knowledge, no method considers the process of state change in engagement.

\textbf{C) Difficulty in annotating engagement:} Users’ engagement is sometimes annotated by annotators based on their subjective (e.g., \cite{del2020you}). However, engagement is not observable due to the user's internal state. In fact, it is reported that engagement annotations vary from annotator to annotator \cite{inoue2019latent}. Because it is difficult to annotate user engagement directly, estimating user engagement using observable features is important.

\textbf{D) Interaction dataset in real-world:} Numerous studies related to engagement estimation conducted experiments in the laboratory. However, using a dataset of interactions observed in a real-world environment is desirable for recording more natural interactions between the robot and users. Although numerous studies use datasets from limited environments, Ben-Youssef et al. have published datasets based on experiments in real public spaces \cite{ben2017ue}.

Some previous studies propose solutions to one of the four open problems. However, when estimating user engagement in public spaces, an estimation method that can simultaneously solve all four open problems is important. Therefore, in this study, we aim to propose an estimating method to solve the four open problems. The contribution of this study is to provide a new engagement estimation framework. In particular, the proposed approach makes two detailed contributions to the engagement estimation framework. First, the proposed model examines the impact of multi-party interactions that can consider the various sizes of different groups. Second, the proposed model represents the continuous process of changes in users engagement from users’ behavior. The proposed model is validated using a dataset of interactions with robots measured in a shopping mall.

\section{Data Collection in Shopping Mall}
\subsection{Experiment Overview}
\begin{figure}[!t]
 \centering  
 \includegraphics[keepaspectratio,width=0.65\linewidth]{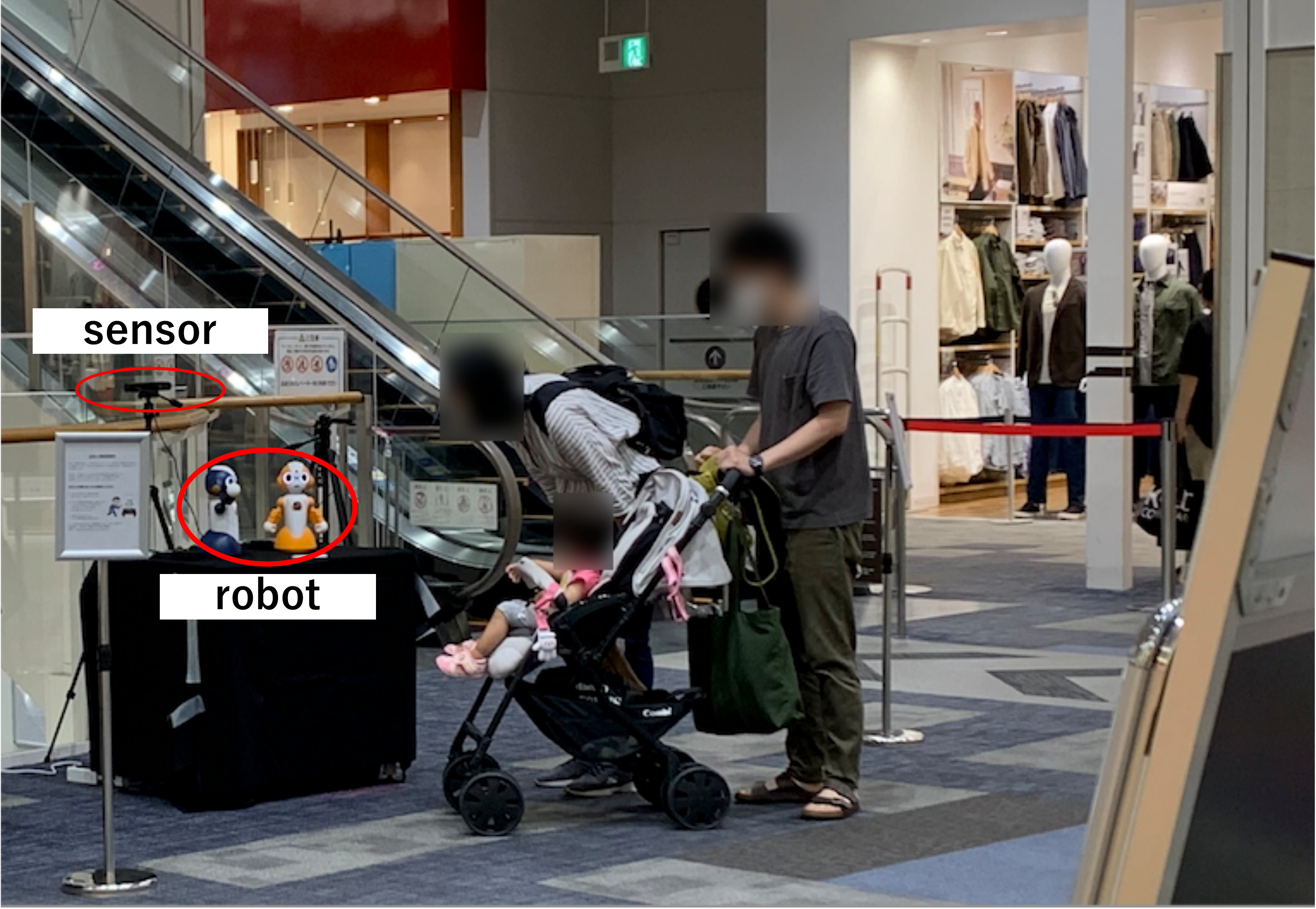}
  \caption{One of the experimental scenes in a shopping mall.}
  \label{experiment1}
\end{figure}

\begin{figure}[!t]
 \centering  
 \includegraphics[keepaspectratio,width=0.8\linewidth]{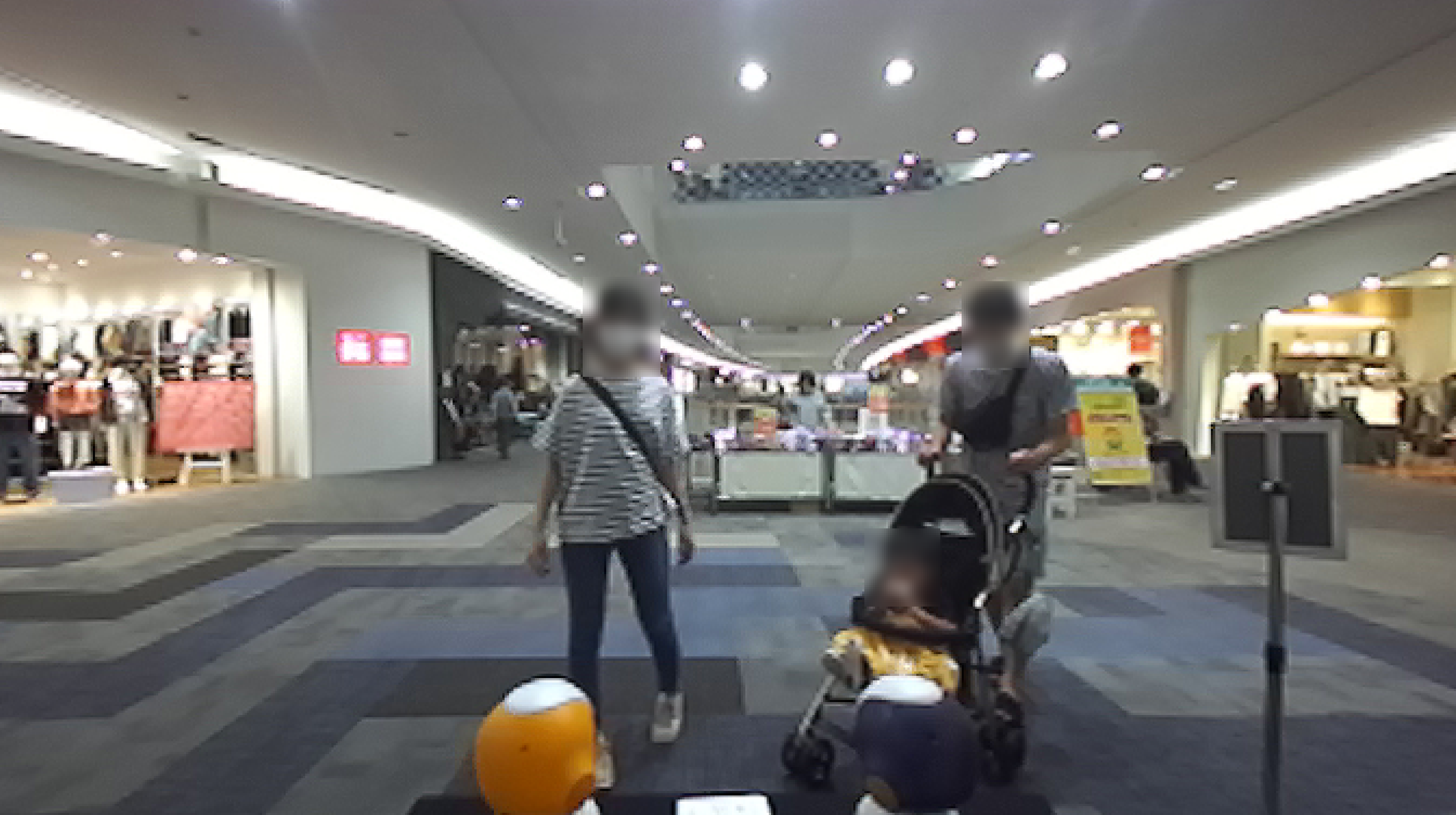}
  \caption{Interaction recorded from the sensor behind the robot.}
  \label{experiment2}
\end{figure}

We conducted an experiment in an actual shopping mall (AEON MALL Kusatsu, Niihama-cho 300, Kusatsu, Shiga, Japan) to record the interactions between social robots and users. The recorded data are the subject of engagement modeling. The experimental setup is illustrated in Fig.~\ref{experiment1}. The participants were all passersby who passed in front of the robots. We recorded the interaction from behind the robot for later analysis, and an example is shown in Fig.~\ref{experiment2}. This experiment was approved by the Research Ethics Committee of the Ritsumeikan University (reference number: BKC-HitoI-2020-027-2).


\subsection{Interaction Design}
The robot ``Sota'' (developed by Vstone Co. Ltd.) was used as the social robot in this experiment. As the interaction design of the robot, we utilize a passive social medium \cite{hayashi2007humanoid} in which two robots talk with each other and do not interact with users actively. Two robots talk with each other at all time and they restart their conversation when the users approach the robots. In other words, the user is simply listens to the robots talk. This type of interaction design has been reported to make it easier for users to be interested in the robot's conversational content itself \cite{hayashi2007humanoid}.

\subsection{Measured Interaction Data}
This experiment was conducted for two days (9am--4:30pm) on weekends in August 2021, and a total of 15 h was measured. From the recorded interactions, only the data of the users who stayed in front of the robot for more than 30 s were used as the targets of engagement estimation. As a result, the number of users measured was 124, and the number of groups was 55 (averaged 2.25 people per a group).

\section{Proposed Method}
\subsection{User Behavior Analysis}
\subsubsection{Definition of User Behavior Categories}
To address the open issue of difficulty in annotating engagement, we utilize user behavior category during interaction with the robot as an observable state for modeling user engagement. This is because user engagement is an internal state of users that we cannot be observed directly. Therefore, we first label the user behavior categories during the interaction that are directly observable and applied for the estimation method in engagement. In some studies, specific human behavior, such as face orientation, is used as input for models such as SVM \cite{xu2013designing, leite2015comparing, salam2015engagement, ben2019fly, inoue2019latent}. However, in this study, we generated a list of behavior categories because of the difficulty to accurately detect the face orientation owing to a noisy environment. We defined user behavior categories from the videos recorded in the experiments. Table~\ref{definition_behavior} lists the user behavior categories which are frequently observed during interactions. The behavior categories and duration of the interaction were annotated from the recorded videos, and they are used as the input for engagement modeling.

Two people unrelated to this study annotated the recorded videos using the ELAN annotation tool. The annotators chose labels from the behavior categories shown in Table~\ref{definition_behavior}. Using the time-sampling method, the label matching rate between the two annotators was 65\%.

\begin{table}[!t]
  \begin{center}
    \caption{Observed user behavior categories}
      \begin{tabular}{c|c} \hline
        behavior &explanation \\ \hline\hline
        \textit{Prowl} & prowling at around the robot \\ \hline
        \textit{Gaze} & gazing at robots  \\ \hline
        \textit{LookAround} & looking around environment except robot \\ \hline
        \multirow{2}{*}{\textit{DoingOthers}} & taking actions not related to interaction \\
         & (e.g., using smartphone) \\ \hline
        \textit{Pointing} & pointing at robot  \\ \hline
        \textit{TalkToRobot} & talking to robot  \\ \hline
        \textit{Touch} & (try to) touching the robot \\ \hline
        \textit{WaveHands} & waving hands to robot  \\ \hline
      \end{tabular}
    \label{definition_behavior}
  \end{center}
\end{table}

\begin{table}[!t]
  \begin{center}
    \caption{Dependence of behavior categories}
      \begin{tabular}{c|c} \hline
        Behavior & with dependence\\ \hline\hline
        \textit{Prowl}       & \checkmark \\ \hline
        \textit{Gaze}        & - \\ \hline
        \textit{LookAround}  & \checkmark \\ \hline
        \textit{DoingOthers} & \checkmark \\ \hline
        \textit{Pointing}    & - \\ \hline
        \textit{TalkToRobot} & - \\ \hline
        \textit{Touch}       & - \\ \hline
        \textit{WaveHands}   & - \\ \hline
      \end{tabular}
    \label{dependent}
  \end{center}
\end{table}

\subsubsection{Dependence of behavior}
\label{action:dependence}
When multiple users interact with the robot simultaneously, one's behavior is influenced by the others' behavior \cite{leite2015comparing, salam2015engagement}. The behavior categories shown in Table~\ref{definition_behavior} represent personal behavior, whereas we additionally consider the influence of others' behavior related to multiparty interactions as the first open problem. We define the influence from others' behavior as ``dependence'' and choose behavior categories with the dependence of behavior from Table~\ref{definition_behavior}. Table~\ref{dependent} shows the defined behavior categories with and without dependence of the behavior.

As an example of the dependence of behavior, we show a concrete example of a \textit{Prowl} behavior that is with dependence. In this experiment, we observed many situations in which parents and children interacted with robots. In this case, children are strongly interested in robots, but parents prowl without interacting with the robot. In this situation, parents are waiting for their children to get bored with the robots, but children do not care about their parents and continue to interact with the robot. Therefore, it can be assumed that the behavior of parents is influenced by that of the children (dependent); however, the  behavior of children is not influenced by that of the parents (independent). Therefore, we determined whether the other behavioral categories were also dependent on other behaviors. The behavioral categories with dependence are also considered to be behaviors that are observed only in the case of multi-party interactions.

\subsection{Proposed Method for Estimating Engagement}
\label{method:modeling}
\begin{figure}[!t]
 \centering  
 \includegraphics[keepaspectratio,width=1.0\linewidth]{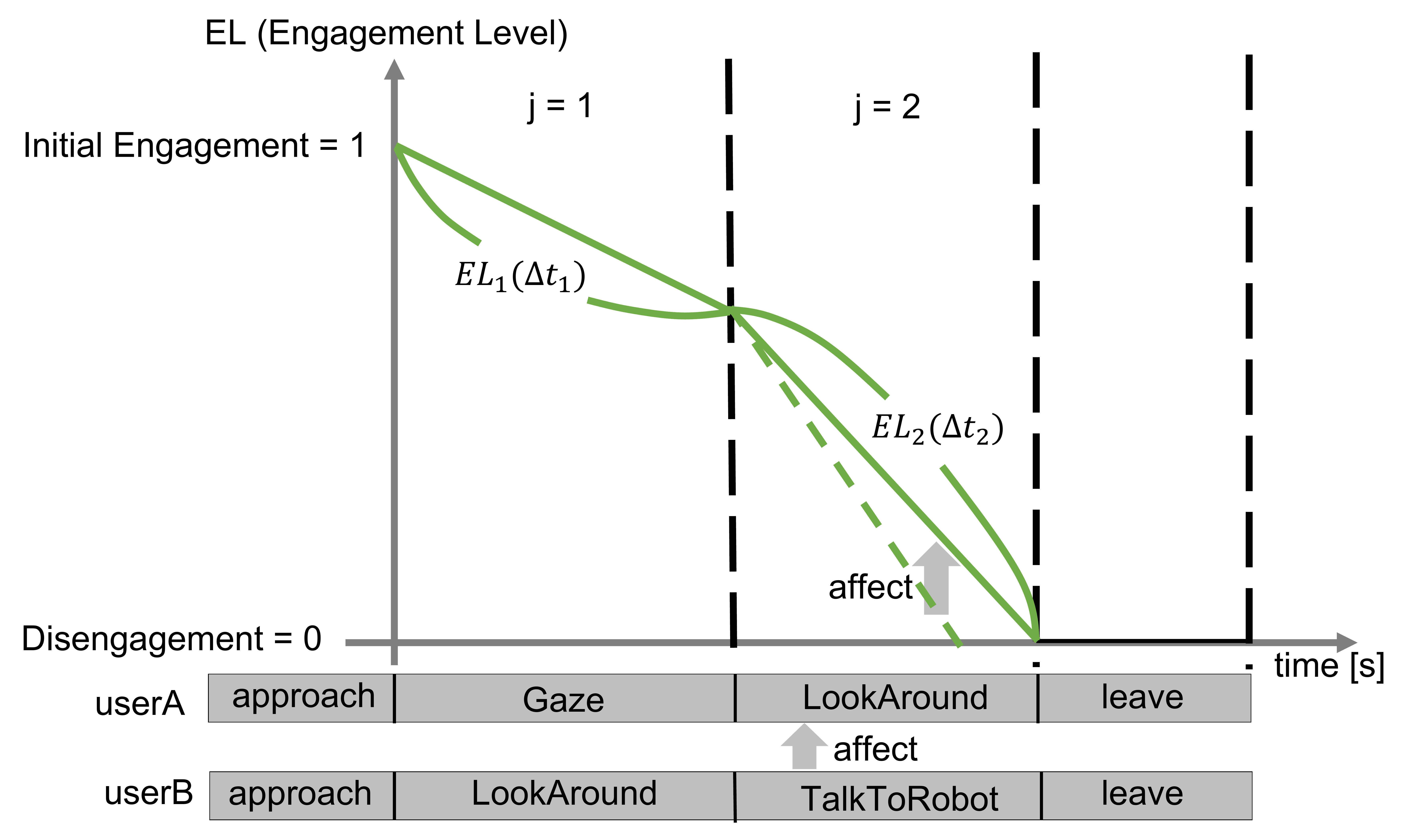}
  \caption{Overview of the proposed method. The change in the engagement value of the User A affected by the behavior of User B is depicted. The slopes of engagement are determined by the behavioral category.}
  \label{method}
\end{figure}

We propose an estimation method for the process of state change in engagement by utilizing observable user behavior categories. In this study, we estimate the rate of changes in engagement from the user's observable behavior because we assume that the user's behavior is determined by engagement representing the user's internal state. An overview of the proposed method is presented in Fig.~\ref{method}. 

First, the proposed method chronologically divides the interacting sections according to the order of behavioral categories observed during the interaction. Thereafter, the rate of change in engagement related to each behavioral category is expressed by (1), defined by a linear model. This represents the process of state change in engagement as the second open problem. We can predict the time when users terminate the interaction with the robot by expressing changes in engagement by (1), .

\begin{align}
  EL_{j}&(\Delta t_j) = a_j\Delta t_j + EL_{j-1}(\Delta t_{j-1}^e)\\
&where\;EL_{j-1}(\Delta t_{j-1}^e)=1 \ (j=1), \notag
\end{align}
where $j$ is the number of the sections which divided by the behavior categories, $\Delta t_j$ denotes the elapsed time and $EL_{j}(\Delta t_j)$ indicates the engagement in each section. Additionally, $EL_{j-1}(\Delta t_{j-1}^e)$ is the last engagement value in the previous section. $a_j$ represents the slope of the engagement, which is determined by the behavior category. The initial value of $EL$ is set 1, and the time when $EL$ firstly becomes 0 is the estimated end time of interaction.

In the proposed method, a probability distribution regarding the slope of the linear model as the rate of change in engagement is assumed for each behavior category. As the example shown in Fig.~\ref{method}, the User A gazes at the robot in Section $j=1$, and the slope of the linear model in this section is sampled from the distribution of the \textit{Gaze} behavior. In this study, we assume normal distribution (2) as the slope of each behavior category. 

\begin{equation}
  a_j \sim p(x_j)=\frac{1}{\sqrt{2\pi\sigma_{x_j}^2}}\exp\left(-\frac{(a_j-\mu_{x_j})^2}{2\sigma_{x_j}^2}\right),
\end{equation}
where $x_j$ is the behavior category observed in section $j$ and $\mu_{x_j}$ and $\sigma_{x_j}$ are the mean and variance of the distribution of behavior $x_j$, respectively.

In terms of the first open problem, during multiparty interaction, the user who is acting the behavior $x_j$ with dependence is influenced by other users' behavior $\bm{y_j}$, which is observed simultaneously. Thus, the dependence of the behavior is expressed by (3), which is multiplying by the probability distributions.
\begin{equation}
  p(x_j|\bm{y_j}) =
 \begin{cases}
   p(x_j) & (independent) \\
   \eta p(x_j)\prod_{s=1}^{S}p(y_{j}^s) &  (dependent)
 \end{cases},
\label{eq:reliance}
\end{equation}
where $\eta$ is a normalizing constant and $S$ is the number of other users who are interacting with the robot simultaneously. 

By representing the dependence of behavior by (3), the slope of each behavior category with dependence becomes close to the slope of the behavior of another user. For instance, in Fig.~\ref{method}, the User A exhibits  \textit{LookAround} behavior in Section $j=2$ because the User A gets bored at robots. Thus, this behavior is considered to have a gradient that reduces his/her engagement (dotted line). Conversely, if the User B behaves as a \textit{TalkToRobot}, which is considered to be of interest to the robot, the slope of \textit{LookAround} is modified to a larger value than the original value (solid line) by (3). As a result, although User A already gets bored of robots, the proposed method estimates that the interaction duration is further extended owing to the influence of User B.

Hereafter, the proposed method 1 is a method in which all behavior categories are assumed to be independent, and the proposed method 2 is a method that considers the dependence of the behavior shown in Table~\ref{dependent}.

\subsection{Training method}
As shown in Section \ref{method:modeling}, our proposed method assumes a normal distribution for the slopes of each behavior category in the change of engagement. Therefore, the mean $\bm{\mu}$ and variance  $\bm{\sigma}$ of the probability distributions for all behavior categories are trained using maximum likelihood estimation. The likelihood function $L$ is designed by (4), which uses the observed interaction duration $t_n^*$ and the estimated interaction duration $\hat{t}_n$ by (1). The likelihood function is optimized by using the quasi-Newton method.

\begin{align}
    L&=-\prod_{n=1}^{N}f(\hat{t}_n;\bm{\mu}, \bm{\sigma}), \\
  f(\hat{t}_n )&=\frac{1}{\sqrt{2\pi\sigma_t^2} }\exp\left(-\frac{(\hat{t}_n-t_n^*)^2}{2\sigma_{t}^2}\right), \\
  \sigma_{t}^2 &= \left( \alpha t_n^* \right)^2,
\end{align}
where $N$ denotes the number of people interacting with the robot to train the proposed method. We assumed that the longer the observed interaction duration, the higher the estimated error. Therefore, the variance $\sigma_t^2$ is calculated using the proportional constant $\alpha$ (in this study, $\alpha = 0.1$) and the observed interaction duration. Additionally, $\hat{t}_n$ is calculated as the time at which the value of engagement represented by (1) becomes $EL = 0$ for the first time. Accordingly, we can estimate engagement using the observable interaction duration without directly annotating engagement.

Of the 124 people whose data were recorded in the shopping mall, 98 people (79\%) were used to train the model, and 26 people (21\%) were used to validate the accuracy of the model. In addition, during training and validation, we compared two proposed methods, which are without and with dependence of behavior.

\section{Evaluation Results}
\subsection{Estimation Results}
\begin{figure}[!t]
 \centering  
 \includegraphics[keepaspectratio,width=0.9\linewidth]{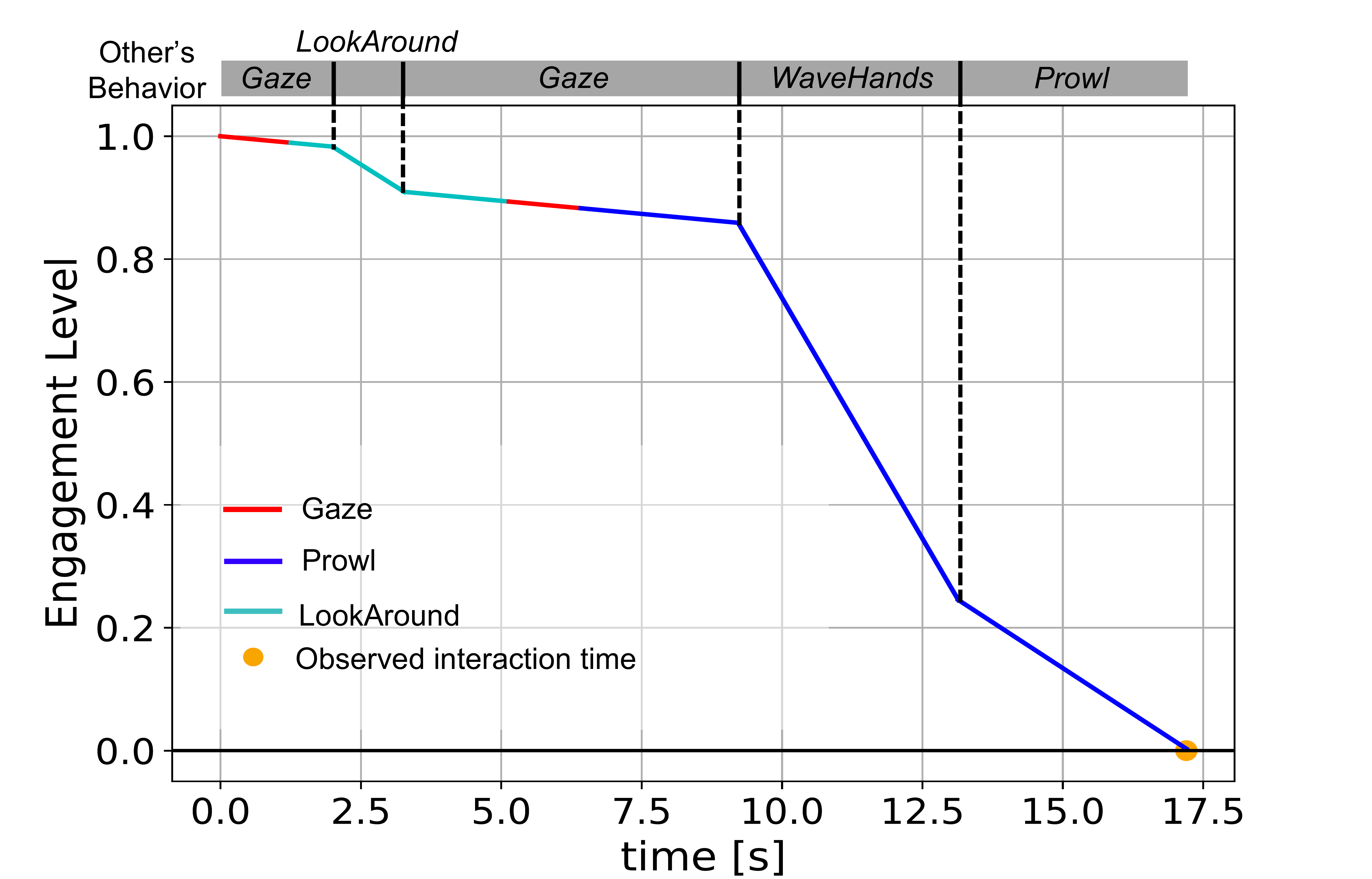}
  \caption{Example of estimation by proposed method 2. The start time of the interaction was 0 s, and the yellow dot represents the observed interaction duration. The color of the engagement indicates the interval in which each behavioral category is observed. The slope of engagement is influenced by other's behavior.}
  \label{exp}
\end{figure}

Fig.~\ref{exp} shows an example of estimation by proposed method 2. In this example, the user shows four behaviors in order of \{\textit{Gaze}, \textit{LookAround}, \textit{Gaze}, and \textit{Prowl}\}, and engagement is changed 	in accordance with these behaviors. The slopes of engagement during \textit{LookAround} and \textit{Prowl} change because other users’ behavior changed during \textit{LookAround} and \textit{Prowl}. Because the estimated interaction duration, which is a time when estimated engagement is zero, is close to the observed interaction duration, the estimation accuracy of the proposed method 2 is high. The estimation accuracy of engagement is also expected to be high due to the high estimation accuracy of the interaction duration.

\begin{figure}[!t]
 \centering  
 \includegraphics[keepaspectratio,width=0.9\linewidth]{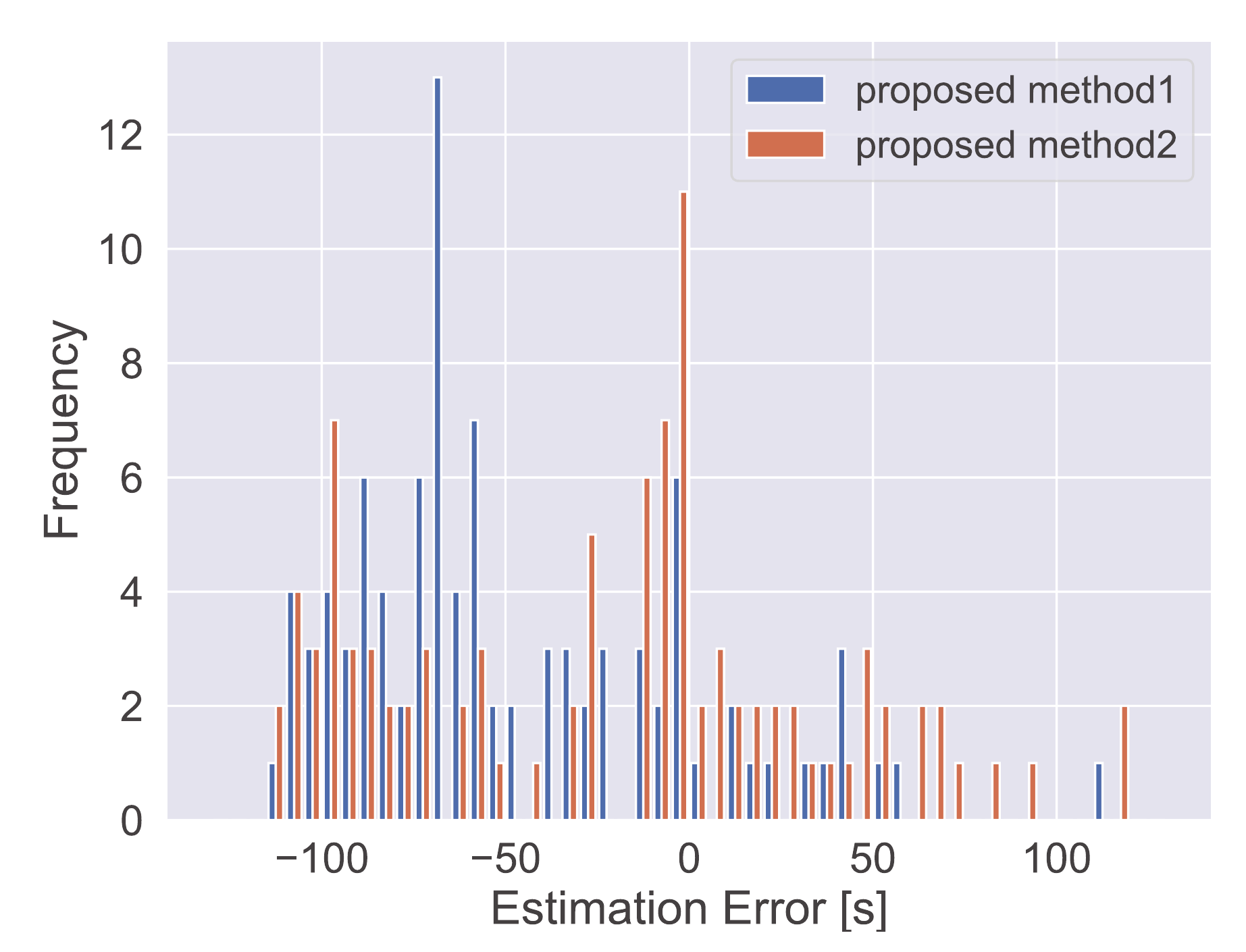}
  \caption{Histogram of estimation error in training which is the difference between the observed and estimated interaction duration.}
  \label{distiribution:train}
\end{figure}

Fig.~\ref{distiribution:train} shows the histogram of the estimation error for each proposed method, which is the difference between the observed and estimated interaction duration, in the training data. Table~\ref{train_error:stats} shows the mean average error (MAE), median, and mode of the estimation error for each method. The closer these values are to zero, the more accurately the proposed method estimates the observed duration of the interaction. The results show that the proposed method 2, which is with the dependence of behavior, can estimate interaction duration more accurately than the proposed method 1 because all three indicators are small.

\begin{figure}[!t]
 \centering  
 \includegraphics[keepaspectratio,width=0.9\linewidth]{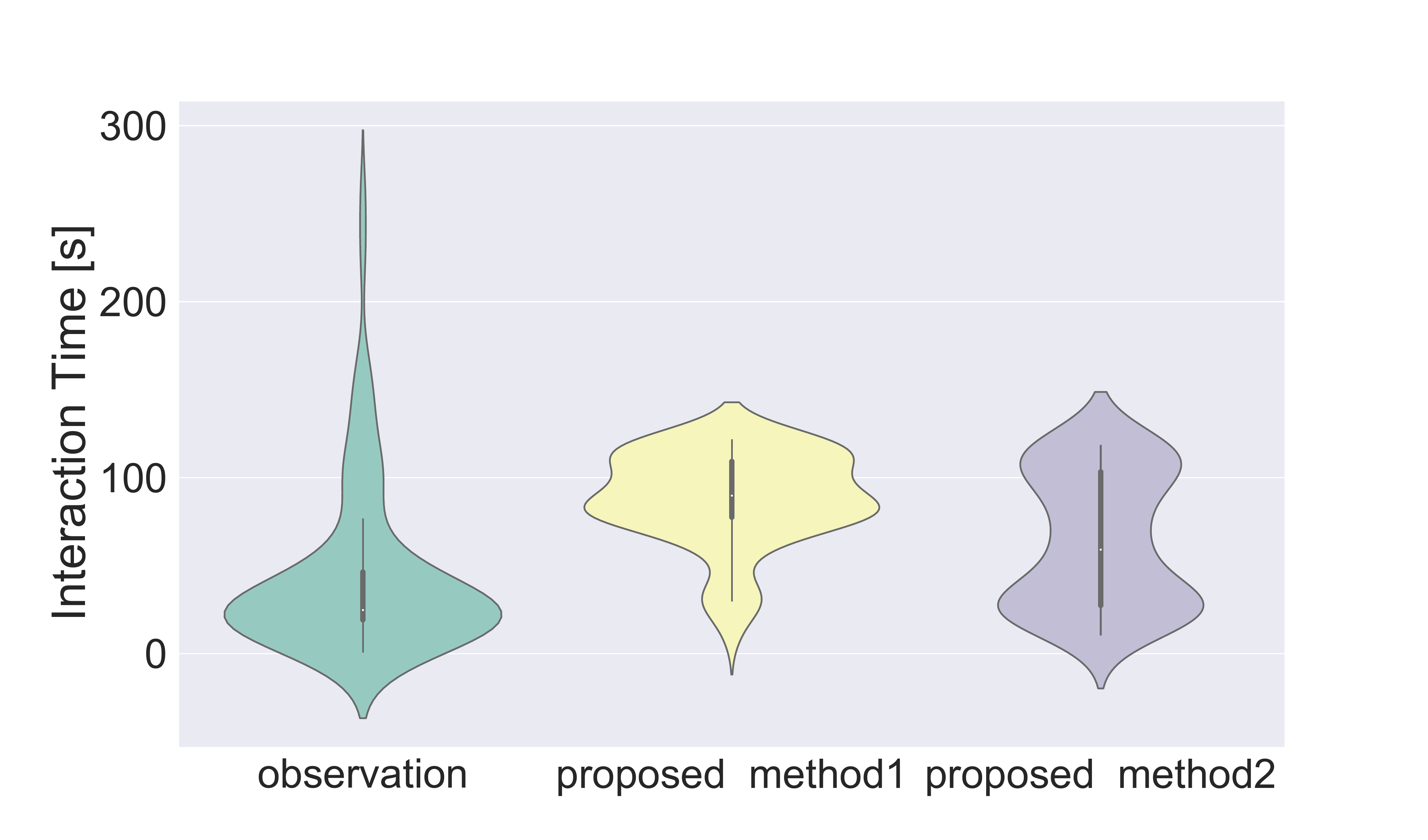}
  \caption{Violin plots of observed and estimated data for each method in training.}
  \label{distiribution}
\end{figure}

Fig.~\ref{distiribution} shows distributions of interaction duration of observed and estimated data for each method in training data. The results show that the peak of the estimated distribution by the proposed method 1 is far from the peak of the observed distribution, whereas the peak of the estimated distribution by the proposed method 2 is closer to the observed distribution. However, the estimated distribution by the proposed method 2 has another peak around 100 s, while the observed data shows a lower density around 100 s.

\begin{table}[!t]
  \centering
    \caption{MAE, median, and mode of estimation error for training}
      \begin{tabular}{c|c|c|c} \hline
        Method & MAE s & median s & mode s \\ \hline\hline
        proposed method 1 & $59.8$ & $-60.9$ & $-65.0$ \\
      proposed method 2 & $50.3$ & $-10.0$ & $0.0$\\ \hline
      \end{tabular}
    \label{train_error:stats}
\end{table}

\begin{figure}[!t]
 \centering  
 \includegraphics[keepaspectratio,width=0.9\linewidth]{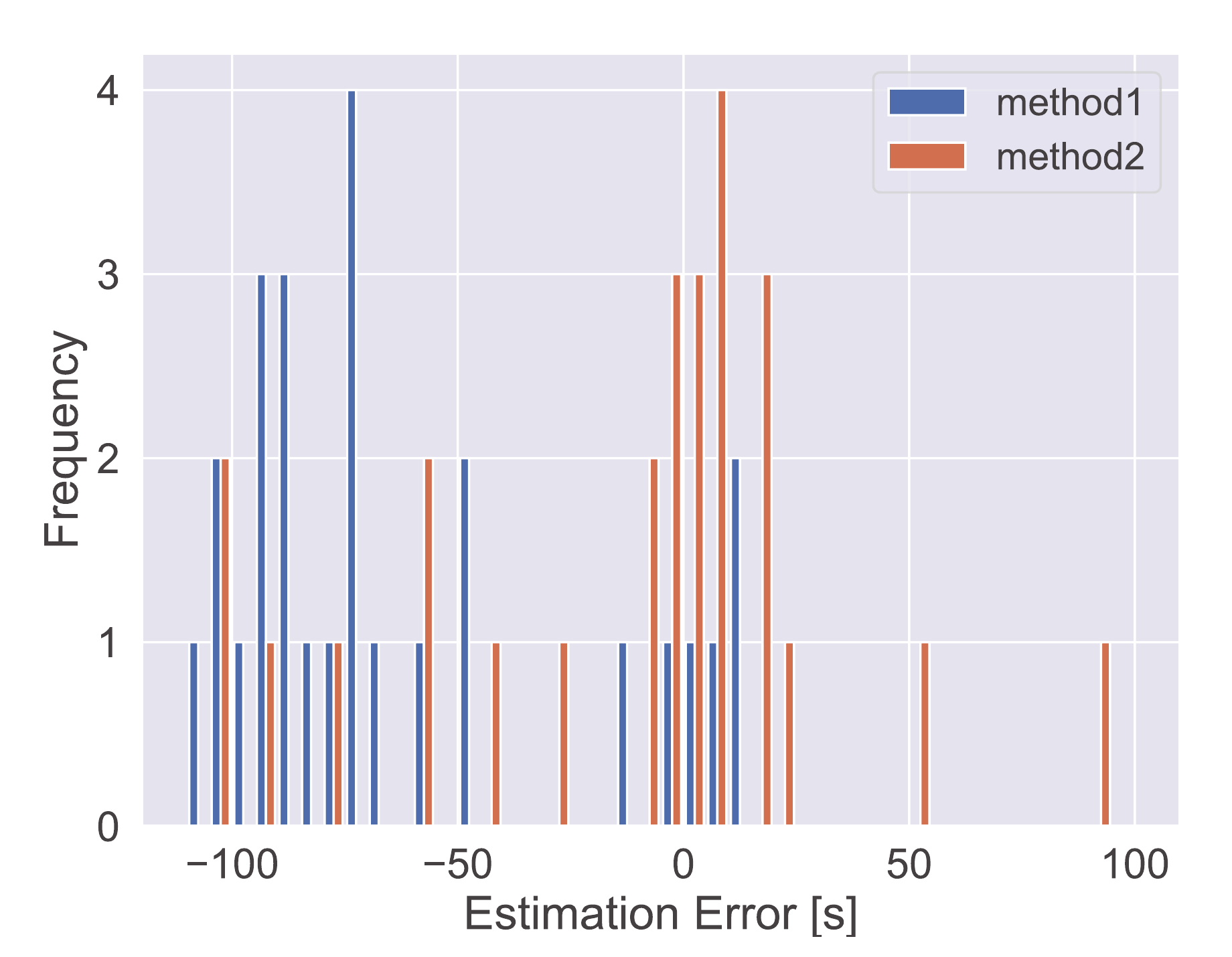}
  \caption{Histogram of estimation error in validation which is the difference between the observed and estimated interaction duration.}
  \label{error_distribution:varidation}
\end{figure}

Next, the accuracy of the proposed methods is verified using validation data from 26 users. Fig.~\ref{error_distribution:varidation} shows a histogram of the estimation error for each method in the validation dataset. Table~\ref{varidation_error:stats} lists the MAE, median, and mode of the estimation error for each method. Even with validation data, the proposed method 2 with dependence of behavior can estimate the interaction duration more accurately than the proposed method 1.

\begin{figure}[!t]
 \centering  
 \includegraphics[keepaspectratio,width=0.9\linewidth]{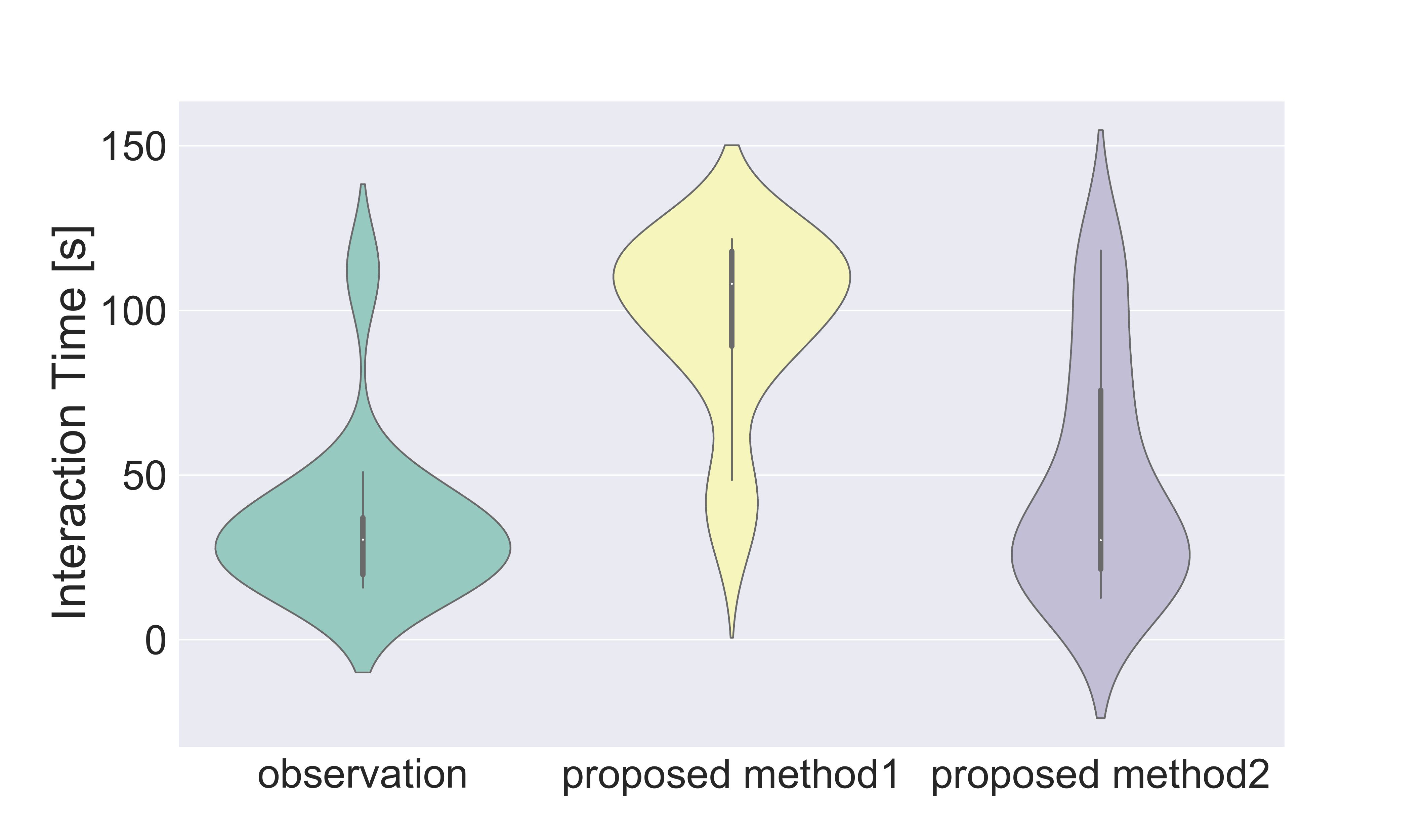}
  \caption{Violin plots of observed and estimated data for each method in validation.}
  \label{distiribution:varidation}
\end{figure}

Fig.~\ref{distiribution:varidation} shows distributions of interaction duration of observed and estimated data for each method in validation data. The results show that the shape of the estimated distribution by proposed method 2 is similar to the shape of observed distribution than the estimated distribution by proposed method 1. Therefore, the proposed method 2 can estimate the interaction duration with higher accuracy than the proposed method 1. 

\begin{table}[!t]
  \centering
    \caption{MAE, median, and mode of estimation error for validation}
      \begin{tabular}{c|c|c|c} \hline
        method & MAE s & median s & mode s \\ \hline\hline
        proposed method 1 & $64.1$ & $-73.3$ & $-70.0$ \\
        proposed method 2 & $32.3$ & $-0.02$ & $10.0$\\ \hline
      \end{tabular}
    \label{varidation_error:stats}
\end{table}

\subsection{Trained probability distribution for behavior categories}
Because the proposed method 2 shows a higher estimation accuracy than the proposed method 1, only the parameters of the distributions for the behavior categories obtained by the proposed method 2 are shown in Table~\ref{result:dependent}. In Table~\ref{result:dependent}, the lower the mean value, the greater the reduction in engagement during the section regarding to its behavior category. We can observe that the frequency of occurrence of each behavior category in the training dataset varies widely. The parameters of each behavior category with a small frequency of occurrence tend not to change much from the initial values $(\mu = -0.15, \sigma = 0.1)$. In particular, the variance means the ambiguity of the parameters; thus, the values of the behaviors that appeared less than 50 times were not trained. Therefore, we consider that only those behavior categories that appeared more than 50 times are correctly trained and compare the parameters of those behaviors. 

The results show that the mean value increases in the order of \textit{Gaze}, \textit{DoingOthers}, \textit{LookAround}, and \textit{Prowl}. Because the mean of \textit{Gaze} is nearly zero, the \textit{Gaze} behavior can be interpreted as behavior that maintains user engagement. Conversely, \textit{DoingOthers}, \textit{LookAround}, and \textit{Prowl} can be interpreted as behavioral categories that decrease user engagement. 

\begin{table}[!t]
  \centering
    \caption{Parameters of probabilistic distribution for each behavior category trained by proposed method 2}
      \begin{tabular}{c|c|c|c} \hline
        behavior category & mean & variance & Frequency \\ \hline\hline
        \textit{Prowl} & $-0.012$ & $0.279\times10^{-6}$ & 290 \\
        \textit{Gaze} & $-0.800\times10^{-2}$ & $0.180\times10^{-2}$ & 271 \\
        \textit{LookAround} & $-0.011$ & $1.000\times10^{-6}$ & 161 \\
        \textit{DoingOthers} & $-0.010$ & $0.347\times10^{-3}$ & 53\\
        \textit{Pointing} & $-0.130$ & $0.095$ & 27 \\
        \textit{TalkToRobot} & $-0.128$ & $0.095$ & 18 \\
        \textit{Touch} & $-0.118$ & $0.115$ & 9 \\
        \textit{WaveHands} & $-0.157$ & $0.093$ & 7 \\ \hline
      \end{tabular}
    \label{result:dependent}
\end{table}

\subsection{Discussion}
From estimation error results in Figs.~\ref{distiribution:train}, \ref{error_distribution:varidation}, and Tables~\ref{train_error:stats}, \ref{varidation_error:stats}, the proposed method 2, which is with dependence of behavior, can estimate the interaction duration more accurately than the proposed method 1. This results indicate that user engagement is influenced by the behavior of other users, consistent with the results of previous studies \cite{leite2015comparing, salam2015engagement}. In particular, the result that the mode of estimation error is approximately zero in  Tables~\ref{train_error:stats}, \ref{varidation_error:stats} implies that many situations of the interactions can be accurately estimated. However, because the MAE of the estimation error for the proposed method 2 is far from zero, its estimation accuracy is not high; this result implies that it cannot be applied in a real environment yet. 

To apply the estimation method to real environments, the authors consider that the estimation error with less more 10\% should be achieved. However, the proposed method cannot satisfy these requirements. We considered three limitations that the proposed method could not estimate the interaction duration sufficiently. First, we used linear models for the process of change in engagement, which is a poor representation of the internal state of the users. Second, the behavior of users is influenced by the behavior of others; however, the proposed model cannot consider the impact of past behaviors of the target user. Finally, owing to the low frequency of the appearance of behaviors, the parameters of all behavior categories could not be accurately trained. A larger dataset was required to train the model accurately. We assume that the proposed estimation method can be achieved with higher accuracy by solving these problems in future work.

\section{Conclusion}
In this study, we proposed a new user engagement estimation framework during interaction with a robot. The proposed model focuses on four unsolved open problems: multiparty interactions, process of state change in engagement, difficulty in annotating engagement, and interaction dataset in the real world. The results show that the proposed model with dependence of behavior can accurately estimate the interaction duration than that of the proposed method without dependence. This implies that the accuracy of engagement estimation improves when the behavior of others is considered. In addition, because the interaction duration can be estimated, these results suggest that the proposed model can also indirectly estimate users’ engagement.

However, because the estimation accuracy is not sufficiently accurate for application in a real-world environment, we need to improve the proposed model by considering three factors: using the nonlinear model to represent engagement, the influence of the target user’s past behavior, and the increase in the amount of data. In addition, Oteral et al. point out that other open questions about estimating engagement such as long-term interactions \cite{oertel2020engagement}. Therefore, a solution to these unsolved points needs to be added to the proposed model. Finally, engagement estimation methods are expected to be applied to flexible robot behavior strategies based on the user's state.

\end{document}